\documentclass{bmvc2k}

\newcommand{\bm}[1]{\mathbf{#1}} 

\newcommand\raiseT[2]{%
\setbox0\hbox{$#1{#2}$}\raise\dp0\box0}

\usepackage{amsmath,amssymb}
\usepackage{booktabs}

\usepackage{graphicx,wrapfig}

\title{Masked Supervised Learning for Semantic Segmentation}

\addauthor{Hasib Zunair}{hasibzunair@gmail.com}{1}
\addauthor{A. Ben Hamza}{hamza@ciise.concordia.ca}{1}

\addinstitution{
 Concordia University \\
 Montreal, QC, Canada
}

\runninghead{Zunair, Hamza}{Masked Supervised Learning for Segmentation}

\begin{document}

\maketitle

\begin{abstract}
Self-attention is of vital importance in semantic segmentation as it enables modeling of long-range context, which translates into improved performance. We argue that it is equally important to model short-range context, especially to tackle cases where not only the regions of interest are small and ambiguous, but also when there exists an imbalance between the semantic classes. To this end, we propose Masked Supervised Learning (MaskSup), an effective single-stage learning paradigm that models both short- and long-range context, capturing the contextual relationships between pixels via random masking. Experimental results demonstrate the competitive performance of MaskSup against strong baselines in both binary and multi-class segmentation tasks on three standard benchmark datasets, particularly at handling ambiguous regions and retaining better segmentation of minority classes with no added inference cost. In addition to segmenting target regions even when large portions of the input are masked, MaskSup is also generic and can be easily integrated into a variety of semantic segmentation methods. We also show that the proposed method is computationally efficient, yielding an improved performance by 10\% on the mean intersection-over-union (mIoU) while requiring $3\times$ less learnable parameters.
\end{abstract}


\section{Introduction}
The basic goal of semantic segmentation, or simply segmentation, is to classify each pixel in an image into one of the pre-defined semantic categories or classes. Its real-world applications are abound, ranging from medical image analysis~\cite{sirinukunwattana2017gland} to robotics~\cite{lai2011large}. In medical imaging, for instance, semantic segmentation can enable physicians to analyze regions of interests (ROIs) more effectively and efficiently for morphological analysis in cancer treatment, especially in high-resolution images~\cite{sirinukunwattana2017gland}, and information retrieval in diagnosis and surgery~\cite{jha2020kvasir}. It also extends to visual scene understanding, which disentangles a scene into objects (e.g. chair), surfaces (e.g. wall) and their relations for robotic object recognition, navigation, manipulation and interaction~\cite{lai2011large}.

\begin{figure}[!htb]
\centering
\includegraphics[scale=.24]{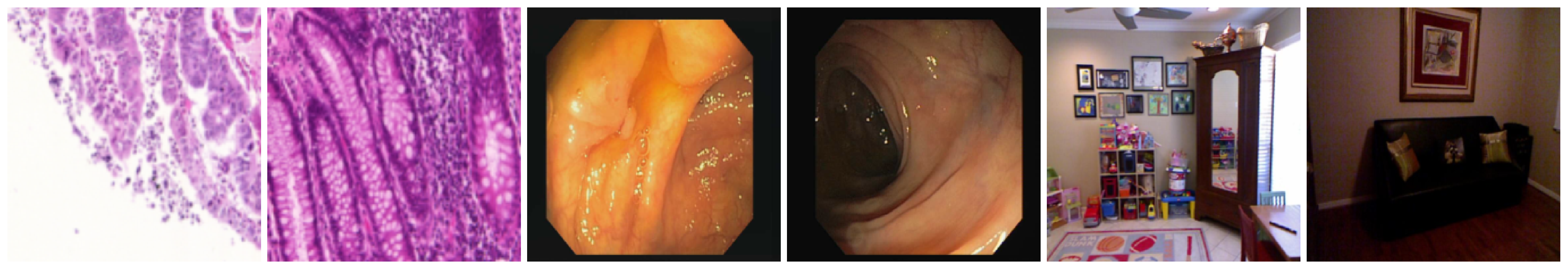}
\caption{Examples demonstrating challenges in semantic segmentation. \textbf{Left to right:} First two from GLaS~\cite{sirinukunwattana2017gland} show the variation in appearance; middle two from CVC-ClinicDB~\cite{bernal2015wm} show difference in scale and ambiguous (ROI); last two from NYU Depth V2~\cite{Silberman:ECCV12} show many classes with heavy imbalance under different lighting conditions.}
\label{Fig:hard_images}
\end{figure}

Previous works on semantic segmentation include FCN~\cite{long2015fully} and U-Net~\cite{ronneberger2015u}, which are convolutional-based encoder-decoder networks, and have been further extended for improved performance~\cite{zunair2021sharp,zhou2019unet++,xiao2018weighted,jha2019resunet++,fang2019selective}. Some recent works have also demonstrated that modeling long-range context, typically via self-attention mechanism~\cite{vaswani2017attention}, translates into better segmentation performance~\cite{oktay2018attention,wang2020axial,valanarasu2021medical,xu2021levit}. Despite the effectiveness of self-attentive models, in this paper we argue that semantic image segmentation is still a challenging problem due to a number of reasons. First, there is diversity in the size and texture of the ROIs~\cite{sirinukunwattana2017gland}. Second, the same type of ROIs may have different sizes and colors due to the label acquisition protocol. Moreover, there are cases of ambiguity, where the boundary between the ROI and the background cannot easily be distinguished~\cite{jha2020kvasir,bernal2015wm}. Third, in the case of natural scenes, there are multiple class instances (i.e. objects and surfaces are cluttered) and there exists imbalance in the semantic classes (e.g. pixels in an image associated with the class \texttt{wall} are more than the class \texttt{chair}), as well as different lighting conditions, making the task much more difficult~\cite{lai2011large}. Examples of these challenging images are shown in Figure~\ref{Fig:hard_images}.

While powerful, most of these self-attentive methods for semantic segmentation tend to over-segment ROIs, output noisy and discontinuous predictions, fail to accurately predict the boundary regions, and poorly segment minority classes. Moreover, they tend to yield misclassification in multi-class image segmentation due in part to the imbalance that exists between the semantic classes and the large number of semantic classes. We argue that the short-range context is equally important to predict small ROIs in medical images, as well as to accurately segment ROIs and reduce misclassification of minority classes in images of natural scenes, where the class instances are dense and cluttered.

To address the above limitations, we propose Masked Supervised Learning (MaskSup), a novel single-stage learning paradigm for semantic segmentation to effectively learn rich and discriminative representations. MaskSup follows a Siamese style network~\cite{bromley1993signature}, where the two branches are identical and share weights. Given an image and its randomly masked version, MaskSup models short-range context among neighboring pixels as the context branch is tasked with predicting the semantic class of masked pixels; thereby leveraging information from non-masked pixels. MaskSup also models global or long-range context by a task similarity constraint, where the similarity of the outputs of the two branches is maximized in order to better learn the shape of class instances, and we find is especially useful in multi-class settings. The main contributions of this work can be summarized as follows:
\begin{itemize}
\item We propose a learning paradigm that aims to model both short- and long-range context via random masking for image segmentation without incurring any additional inference cost.
\item We show through experimental results and ablation studies for binary and multi-class semantic segmentation tasks on three public datasets that MaskSup yields competitive performance in comparison with single and multi-task learning baselines.
\item We demonstrate that MaskSup is robust to large masked corruptions and is computationally efficient, especially in multi-class segmentation as it improves by over 10\% mIoU while at the same time requires $3\times$ less learnable parameters.
\end{itemize}

\section{Related work}
\medskip\noindent\textbf{Single-task semantic segmentation.}\quad Most state-of-the-art segmentation methods usually follow an encoder-decoder network structure, where the image is first downsampled by the encoder subnetwork to a latent representation and then the decoder subnetwork is used to semantically project the latent representation into a pixel space for precise localization. Convolutional-based methods include FCN~\cite{long2015fully} and U-Net~\cite{ronneberger2015u}, and their variants such as U-Net++\cite{zhou2019unet++}, ResU-Net~\cite{xiao2018weighted}, ResU-Net++~\cite{jha2019resunet++}. Due to the inability of convolutional-based methods to model long-range context~\cite{wang2020axial}, self-attention~\cite{vaswani2017attention} has become a core building block in various attention-based methods such as Attention U-Net~\cite{oktay2018attention} and Axial Attention U-Net~\cite{wang2020axial}. To better segment ROIs at boundaries, Selective Feature Aggregation (SFA)~\cite{fang2019selective} employs area-boundary constraints for polyp segmentation. KiU-Net~\cite{valanarasu2021kiu} leverage overcomplete convolutional architectures to better segment very small ROIs and distinguish between ROI and background accurately. More recently, the advent of Vision Transformers~\cite{dosovitskiy2020vit} has accelerated research in the direction of transformer-based segmentation methods, which also build upon self-attention~\cite{vaswani2017attention}. These transformer-based methods include MedT~\cite{valanarasu2021medical} and LeViT-UNet~\cite{xu2021levit}, which aim to learn long-range context. Our proposed framework differs from previous work in that it captures both short- and long-range context, while learning fewer parameters without compromising performance. In fact, masking enables the base segmentation model to learn short-range context among nearby pixels, as the model is tasked to make a pixel-level prediction for a masked input. This forces the network to leverage information from the nearby pixels in order to make a prediction.

\medskip\noindent\textbf{Multi-task semantic segmentation.}\quad Semantic segmentation can be jointly optimized with other visual scene understanding tasks such as depth estimation and edge detection. HybridNet A2~\cite{lin2019depth} is a multi-task learning method, which employs a hybrid convolutional neural network to jointly tackle the task of image segmentation and depth estimation using a single network. PAD-Net~\cite{xu2018pad} is a multi-task learning and distillation based network, which jointly predicts a segmentation, depth, surface normal and edge map by multi-modal data fusion. This is extended in MTI-Net~\cite{vandenhende2020mti}, where interactions between segmentation and depth estimation are captured at multiple scales when distilling information based on multi-modal distillation, in which the tasks mutually benefit from each other. Unlike multi-task learning methods, our method does not require additional training data, and hence reduces the need for intense manual labeling of additional data.

\section{Proposed Method}
We consider the problem of learning an encoder-decoder network $\bm{f}_{\theta}$ that classifies each pixel of an image $\bm{I}$ into its semantic class category. The output is an image $\bm{M}_\text{p} = \bm{f}_{\theta}(\bm{I})$ of the same size as the input image. A gland segmentation task, for example, can be thought of as a binary segmentation problem with two semantic classes: \texttt{gland} and \texttt{background}.

\subsection{Masked Supervised Learning}
We present a masked supervised learning framework for effectively learning rich and discriminative representations for semantic segmentation. The proposed MaskSup method follows a Siamese network~\cite{bromley1993signature} style architecture, in which the segmentation branch (SB) and context branch (CB) are identical and share weights. Given an image $\bm{I}$ and its randomly masked version $\bm{I}_\text{masked}$, we first employ a base segmentation network $\bm{f}_{\theta}$ to output the predictions $\bm{M}_\text{p}$ and $\bm{M}_\text{pm}$, respectively, followed by computing an overall loss function. We use a loss function $\mathcal{L}_\text{context}$ to learn short-range context among neighboring pixels, as this branch \textbf{predicts the semantic class of masked pixels} by leveraging pixel information from the non-masked parts in $\bm{I}_\text{masked}$. We also use a loss term $\mathcal{L}_\text{tasksim}$ to learn long-range context, as the similarity of the two outputs $\bm{M}_\text{p}$ and $\bm{M}_\text{pm}$ is maximized, and hence enables us to better learn the shape of the semantic classes. Overall, MaskSup enables better representation learning for semantic segmentation by capturing the contextual relationships between pixels by predicting a segmentation map for the masked version of the input. MaskSup can tackle cases where the ROIs are ambiguous at boundary, various scales, shape, appearance and also for images that have multiple class instances with imbalance, resulting in fewer pixel-level misclassification of minority classes.

\begin{figure*}[!htb]
\centering
\includegraphics[scale=.8]{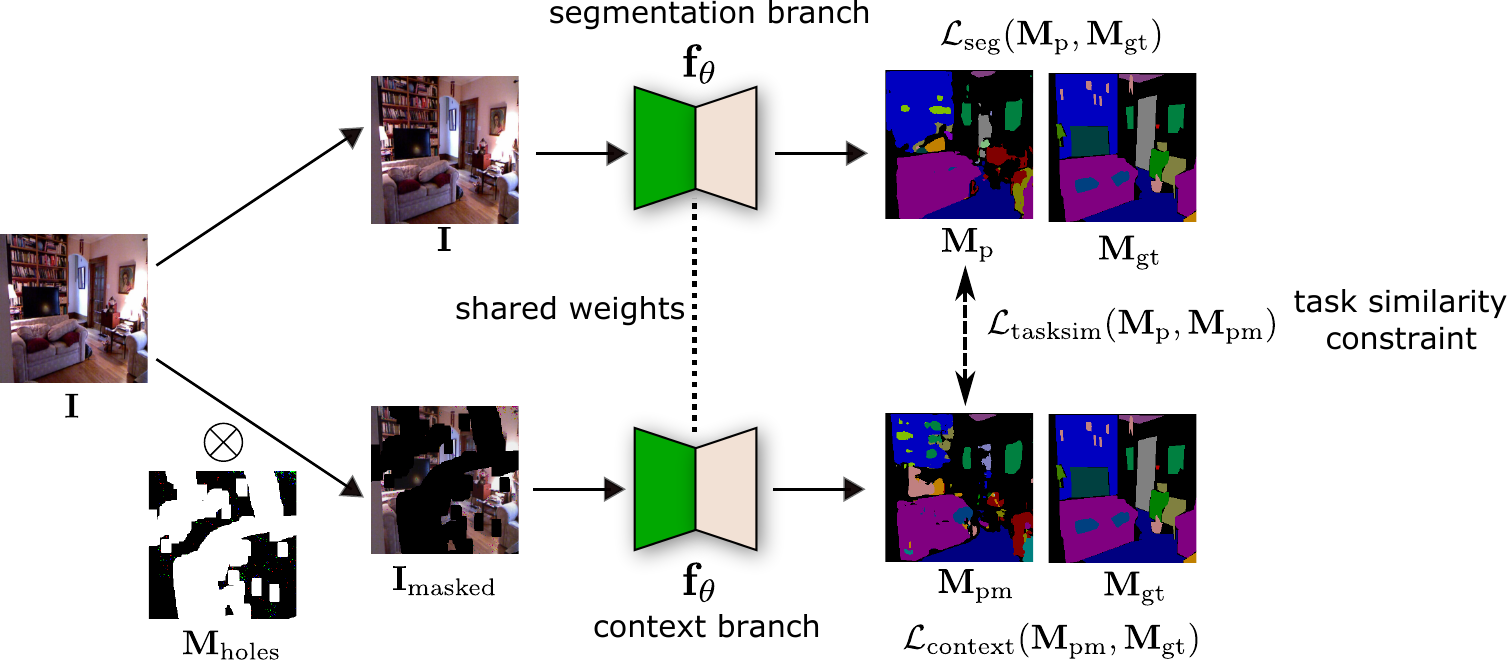}
\caption{\textbf{Overview of MaskSup training}: joint prediction architecture with context branch and task similarity constraint for semantic segmentation, where $\bm{f}_{\theta}$ is a base segmentation network. The segmentation and context branches are identical and share weights.}
\label{Fig:overview}
\end{figure*}

After training, we employ the network $\bm{f}_{\theta}$ to infer a new image $\bm{I}$ that outputs a prediction $\bm{M}_\text{p}$. It is important to mention that at test time the images are not masked, and random masking is only used during training. The overall framework is illustrated in Figure~\ref{Fig:overview}.

\medskip\noindent\textbf{Context Branch.}\quad The goal of the context branch (CB) is to learn short-range context among nearby pixels as this branch outputs predictions for masked pixels by leveraging information from the non-masked pixels in $\bm{I}_\text{masked}$. We take inspiration from the idea of image inpainting, which refers to the task of filling holes in an image, and is commonly used in image editing in order to remove image content such as text in videos or objects~\cite{liu2018image,suvorov2022resolution}. We follow the masking procedure in~\cite{liu2018image} to construct a masked image $\bm{I}_\text{masked}$ from $\bm{I}$ using masks of random streaks and holes of arbitrary shapes
\begin{equation}
\bm{I}_\text{masked} = \bm{I} \odot \bm{M}_\text{holes}
\label{eq:masked}
\end{equation}
where $\odot$ denotes element-wise multiplication and $\bm{M}_\text{holes}$ is a binary mask of random streaks and holes of arbitrary shapes. Intuitively, $\bm{I}_\text{masked}$ has a similar layout as $\bm{I}$, but randomly removes (i.e. pixel values set to 0) roughly over 50\% of the pixels in the image.

Given an input image $\bm{I}$ and the masked image $\bm{I}_\text{masked}$, we train an encoder-decoder network $\bm{f}_{\theta}$ to predict the output of the segmentation branch $\bm{M}_\text{p}$ and the context branch (CB) $\bm{M}_\text{pm}$. Note that $\bm{f}_{\theta}$ is a Siamese network~\cite{bromley1993signature} like architecture where the branches are identical and share weights. In most of our experiments, $\bm{f}_{\theta}$ is either a LeViT-UNet-384~\cite{xu2021levit} for gland and polyp segmentation or U-Net++~\cite{zhou2019unet++} for indoor scene segmentation. We train $\bm{f}_{\theta}$ by minimizing the cross-entropy loss of both the segmentation branch and the context branch for all samples in the training set. The context branch loss is given by
\begin{equation}
\mathcal{L}_\text{CB} = \mathcal{L}_\text{seg} (\bm{M}_\text{p}, \bm{M}_\text{gt}) + \mathcal{L}_\text{context} (\bm{M}_\text{pm}, \bm{M}_\text{gt})
\label{eq:loss1}
\end{equation}
where $\mathcal{L}_\text{seg}$ and $\mathcal{L}_\text{context}$ are cross-entropy losses between the output and ground truth. It is worth mentioning that other application-specific loss functions such as the focal Tversky loss~\cite{abraham2019novel} and distance-based losses~\cite{karimi2019reducing, caliva2019distance} can also be used.

\medskip\noindent\textbf{Task Similarity Constraint.}\quad The goal of the task similarity constraint is to model long-range context by maximizing the similarity between the output from the segmentation branch and the output from context branch in order to predict the semantic classes of masked pixels; thereby enabling us to better learn the shape of the class instances (e.g. \texttt{gland}, \texttt{polyp}, \texttt{wall} etc.). More specifically, we aim to maximize the similarity between the predictions made by the segmentation branch output $\bm{M}_\text{p}$ and the context branch output $\bm{M}_\text{pm}$ by minimizing the $L_2$ error $\mathcal{L}_\text{tasksim}=\Vert\bm{M}_\text{p}-\bm{M}_\text{pm}\Vert_{2}$. Therefore, the overall loss function of MaskSup is a weighted sum of the segmentation, context and task similarity loss terms
\begin{equation}
\mathcal{L}_\text{total} = \alpha_{1} \mathcal{L}_\text{seg} (\bm{M}_\text{p}, \bm{M}_\text{gt}) + \alpha_{2} \mathcal{L}_\text{context} (\bm{M}_\text{pm}, \bm{M}_\text{gt}) + \alpha_{3} \mathcal{L}_\text{tasksim} (\bm{M}_\text{p}, \bm{M}_\text{pm})
\label{eq:loss2}
\end{equation}
where $\alpha_{1}$, $\alpha_{2}$ and $\alpha_{3}$ are nonnegative regularization hyper-parameters, which control the contribution of each loss term. In our experiments, we empirically set them to 1.

During training, the total loss $\mathcal{L}_\text{total}$ is minimized for several epochs to learn the parameters of $\bm{f}_{\theta}$ using a labeled training set $\mathcal{{D}}=\{(\bm{I}_\text{1},\bm{M}_\text{1}),\dots,(\bm{I}_\text{n},\bm{M}_\text{n})\}$, where $\bm{M}_\text{i}$ is the ground truth segmentation mask of the input image $\bm{I}_i$. During testing, the network $\bm{f}_{\theta}$ is used for semantic segmentation, outputting a segmentation mask prediction $\bm{M}_\text{p}$ given an input image $\bm{I}$. The architecture and the different loss terms of MaskSup are illustrated in Figure~\ref{Fig:overview}.

\section{Experiments}
In this section, we present our experimental setup and results in comparison with competing single and multi-task learning baselines for semantic segmentation. Details on datasets, implementation, architecture, training, and additional results are included in the supplementary material. Code is available at: \textcolor{blue}{https://github.com/hasibzunair/masksup-segmentation}

\subsection{Experimental Setup}
\noindent\textbf{Datasets.}\quad We demonstrate and analyze the performance of our method on Gland Segmentation (GLaS)~\cite{sirinukunwattana2017gland}, Kvasir~\cite{jha2020kvasir} \& CVC-ClinicDB~\cite{bernal2015wm} and NYUDv2~\cite{Silberman:ECCV12} datasets. While GLaS and Kvasir \& CVC-ClinicDB are for medical image segmentation tasks, NYUDv2 is for indoor scene segmentation tasks. These datasets cover a wide range of challenges in semantic segmentation, and they represent both binary and multi-class segmentation. They also cover both natural and medical image modalities. In medical images, ROIs are usually very small compared to background. In addition, these datasets have their own challenges such as variation in appearance, scale, ambiguous ROIs, and many class instances with imbalance that are densely cluttered.

\medskip\noindent\textbf{Baselines.}\quad We evaluate the performance of our method against several state-of-the-art convolutional-based methods including FCN~\cite{long2015fully}, U-Net~\cite{ronneberger2015u}, U-Net++\cite{zhou2019unet++}, ResU-Net~\cite{xiao2018weighted}, ResU-Net++~\cite{jha2019resunet++}, SFA~\cite{fang2019selective}, KiU-Net~\cite{valanarasu2021kiu} and attention-based methods such as Attention U-Net~\cite{oktay2018attention}, Axial Attention U-Net~\cite{wang2020axial}. We also compare with more recent transformer-based methods MedT~\cite{valanarasu2021medical} and LeViT-UNet~\cite{xu2021levit}, and multi-task learning methods HybridNet A2~\cite{lin2019depth} and PAD-Net~\cite{xu2018pad} and MTI-Net~\cite{vandenhende2020mti} with HRNet-18~\cite{wang2020deep} as backbone.

\medskip\noindent\textbf{Evaluation Metric.}\quad We report results using the Mean Intersection-Over-Union (mIoU), which is a commonly used metric in semantic segmentation~\cite{ronneberger2015u,wang2020axial,valanarasu2021medical,valanarasu2021kiu,xu2021levit}. The values of mIoU range from 0 to 1, with 1 indicating perfect match between the true and predicted labels, while 0 indicates a complete mismatch between them.

\subsection{Results}
\noindent\textbf{Comparison with State-Of-The-Art.}\quad We compare the performance of MaskSup against several state-of-the-art methods and report the results in Table~\ref{Tab:sota}. All mIoU scores are averaged over 3 runs. As can be seen, MaskSup consistently outperforms all baselines, achieving relative improvements of 1.26\%, 3.45\% and 4.85\% over the strongest baseline in terms of mIoU on GLaS, Kvasir \& CVC-ClinicDB and NYUDv2 datasets, respectively.

\begin{table}[!htb]
\caption{Performance comparison of MaskSup and baselines on GLaS, Kvasir \& CVC-ClinicDB and NYUDv2 test sets using mIoU. Boldface numbers
indicate the best performance, whereas the best baselines are underlined. $\bigtriangleup$ indicates a multi-task learning method.}
\setlength{\tabcolsep}{3.5pt}
\medskip
\centering
\resizebox{0.98\textwidth}{!}{%
\begin{tabular}{l*{7}{c}}
\toprule
\textbf{Method} & \textbf{GLaS, mIoU ($\uparrow$)} & \textbf{CVC-Clinic-DB, mIoU ($\uparrow$)} & \textbf{NYUDv2 ($\uparrow$)} \\
\midrule
U-Net~\cite{ronneberger2015u} & 67.41 & 69.74 & 33.60 \\
FCN~\cite{long2015fully} & 50.84 & - & 29.20 \\
U-Net++\cite{zhou2019unet++} & 69.10 & 72.90 & 34.74 \\
HRNet-18~\cite{wang2020deep} & - & - & 33.18 \\
ResU-Net~\cite{xiao2018weighted} & 65.95 & - & - \\
ResU-Net++~\cite{jha2019resunet++} & - & 79.60 & - \\
SFA~\cite{fang2019selective} & - & 60.70 & - \\
Attention U-Net~\cite{oktay2018attention} & - & 82.70 & - \\
Axial Attention U-Net~\cite{wang2020axial} & 63.03 & - & - \\
MedT~\cite{valanarasu2021medical} & 69.61 & - & - \\
KiU-Net~\cite{valanarasu2021kiu} & 72.78 & - & - \\
LeViT-UNet-128~\cite{xu2021levit} & 70.45 & - & - \\
LeViT-UNet-192~\cite{xu2021levit} & 71.83 & 79.16 & - \\
LeViT-UNet-384~\cite{xu2021levit} & \underline{73.88} & \underline{81.38} & - \\
PAD-Net~\cite{xu2018pad} $\bigtriangleup$ & - & - & 33.10 \\
HybridNet A2~\cite{lin2019depth} $\bigtriangleup$ & - & - & 34.30 \\
MTI-Net~\cite{vandenhende2020mti} $\bigtriangleup$ & - & - & \underline{37.49} \\ \midrule
MaskSup (\textbf{Ours}) & \textbf{76.06} & \textbf{84.02} & \textbf{39.31} \\
\bottomrule
\end{tabular}%
}
\label{Tab:sota}
\end{table}

MaskSup yields significant relative improvements of 18.7\% over Axial Attention U-Net~\cite{wang2020axial} and 2.8\% over KiU-Net~\cite{valanarasu2021kiu} on GLaS. MaskSup also outperforms transformer-based methods such as MedT~\cite{valanarasu2021medical} and LeViT-UNets~\cite{xu2021levit} with relative improvements of 1.26\% and 3.45\% over LeViT-UNet-384~\cite{xu2021levit} on GLaS and Kvasir \& CVC-ClinicDB, respectively. MaskSup performs better than multi-task learning methods PAD-Net~\cite{xu2018pad} and HybridNet A2~\cite{lin2019depth} with relative improvements of 18.76\% and 14.6\%. In addition, MaskSup outperforms MTI-Net~\cite{vandenhende2020mti} with a relative improvement of 4.85\% on NYUDv2. This improvement is significant because MTI-Net~\cite{vandenhende2020mti} is a multi-task learning method that jointly learns four different tasks (i.e. semantic segmentation, depth estimation, edge detection and surface normal estimation), and hence requires additional annotated data for training. In contrast, MaskSup only requires images and the pixel level annotations (i.e. segmentation masks); thereby reducing the need for intense manual labeling of additional data. The results demonstrate the effectiveness and capability of MaskSup in modeling short- and long-range context, yielding improved segmentation.

In Table~\ref{Tab:mae}, we report the performance comparison results of MaskSup and masked autoencoders (MAE)~\cite{he2022masked}. For MAE, we pre-train for 800 epochs on the images and fine-tune for 50 epochs on images and labels, while MaskSup only requires a single stage of training for 200 epochs. MAE uses patch-based masking to predict visual tokens similar to image inpainting, whereas MaskSup outputs a prediction label and not the full inpainted image.
\begin{table}[!htb]
\caption{Performance comparison of MaskSup and MAE. MaskSup is efficient and achieves better performance.}
\medskip
\centering
\begin{tabular}{l*{7}{c}}
\toprule
\textbf{Method} & \textbf{GLaS, mIoU ($\uparrow$)} & \textbf{CVC-Clinic-DB, mIoU ($\uparrow$)} & \textbf{NYUDv2 ($\uparrow$)} \\
\midrule
MAE~\cite{he2022masked} & 75.04 & 82.50 & 37.42 \\
MaskSup (\textbf{Ours}) & \textbf{76.06} & \textbf{84.02} & \textbf{39.31} \\
\bottomrule
\end{tabular}
\label{Tab:mae}
\end{table}

\begin{figure}[!htb]
\centering
\includegraphics[scale=.2]{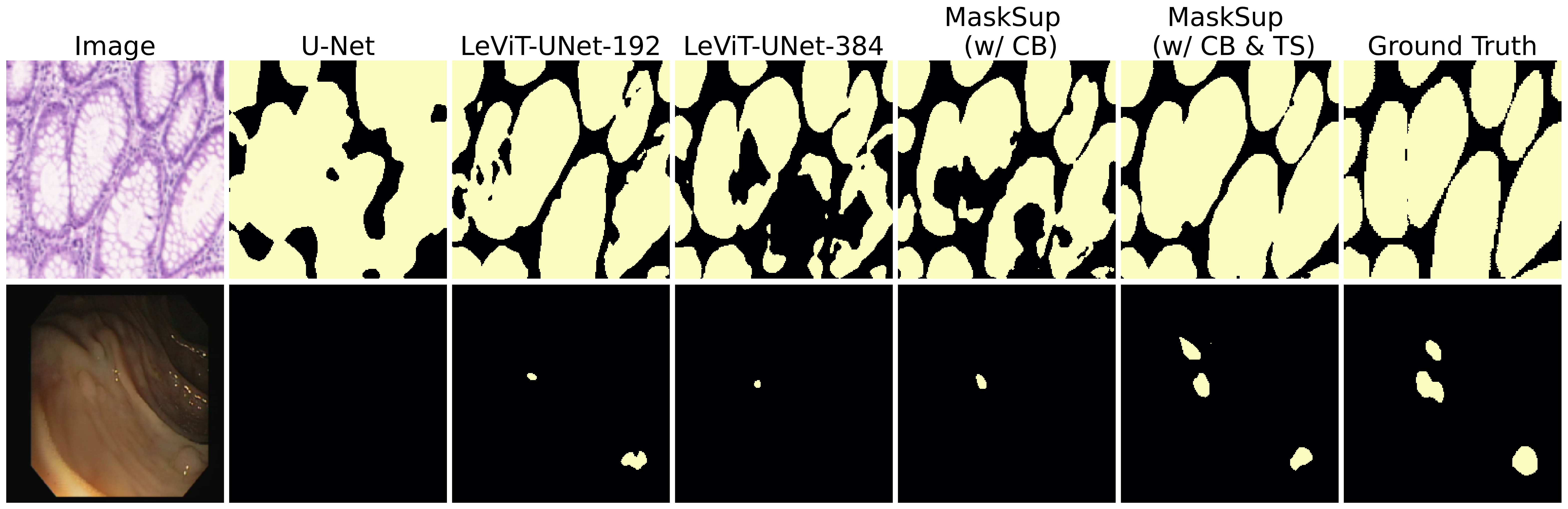}
\caption{Visual comparison of MaskSup and baselines on the GLaS and Kvasir \& CVC-ClinicDB test sets. MaskSup outputs better predictions in cases of variation in overall appearance and also very small and ambiguous ROIs.}
\label{Fig:qual1}
\end{figure}

\begin{figure}[!htb]
\centering
\includegraphics[scale=.23]{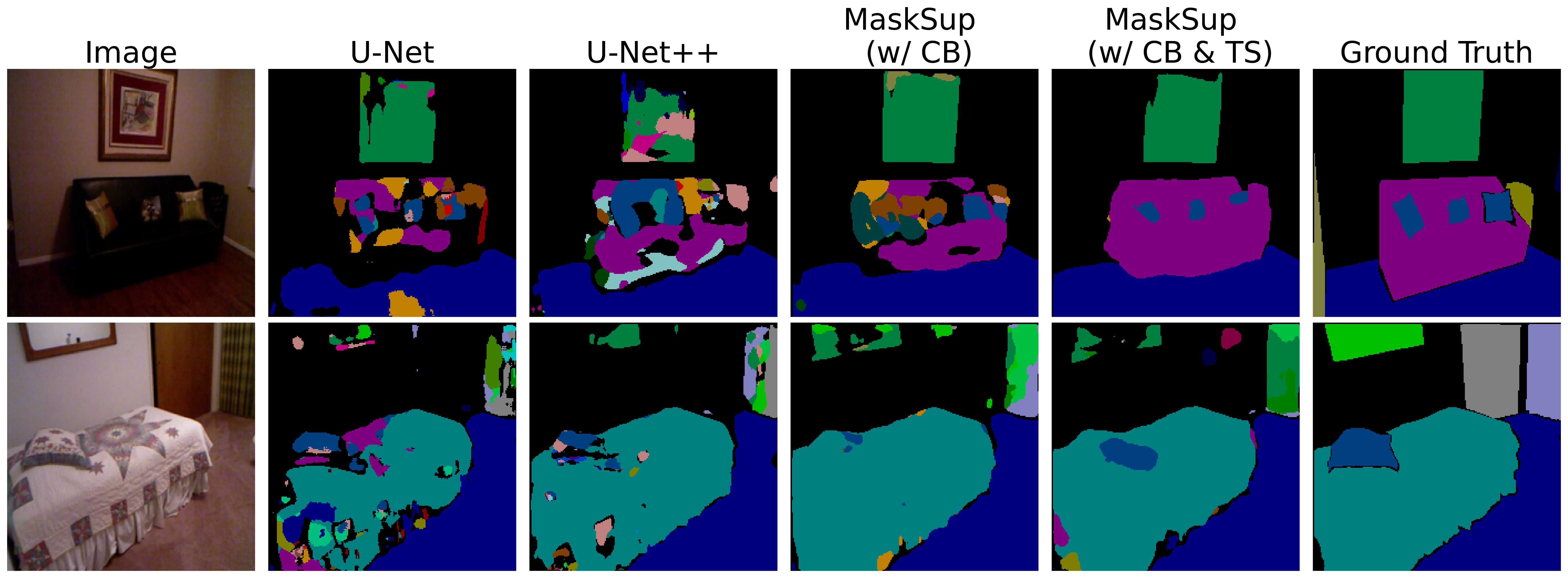}
\caption{Visual comparison of MaskSup and baselines on the NYU Depth V2 test set. MaskSup is able to output better predictions for minority classes (e.g. pillows).}
\label{Fig:qual2}
\end{figure}

\medskip\noindent\textbf{Qualitative Results.}\quad In Figures~\ref{Fig:qual1} and~\ref{Fig:qual2}, we visually compare MaskSup predictions against the baselines U-Net~\cite{ronneberger2015u}, U-Net++\cite{zhou2019unet++} and LeViT-UNets~\cite{xu2021levit} on GLaS, Kvasir \& CVC-ClinicDB and NYUDv2. In the first row of Figure~\ref{Fig:qual1}, when there is a change in the overall shape and appearance of the glands, the baseline methods tend to over-segment the regions and also produce noisy outputs as they fail to capture the global structure and semantics of the glands. The second row of Figure~\ref{Fig:qual1} shows the case of ambiguous ROIs of polyps, where the baselines fail to accurately segment the ROI. This is largely attributed to the limited capability of the learned representations used in the baselines. Interestingly the LeViT-UNets baseline fail to segment ROIs that are ambiguous at boundaries and vary in size and color, albeit transformers are quite strong in modeling long-range context~\cite{dosovitskiy2020vit,wang2020axial}.  Self-attention can be regarded as a form of the non-local means~\cite{buades2005non}, and it captures long-range dependencies, resulting in over-segmentation as shown in Figure~\ref{Fig:qual1}.

Figure~\ref{Fig:qual2} shows that the baselines fail to accurately segment multiple class instances, output discontinuous predictions and misclassify instances. In the last row of Figure~\ref{Fig:qual2}, we can see that the baselines fail to segment the minority class (i.e. pillow). Overall, the baselines fail to capture context of target regions, resulting in over-segmentation, noisy and discontinuous predictions as well as misclassification of instances, leading to unsatisfactory predictions.

By comparison, MaskSup is able to better capture the shape and appearance of instances due, in large part, to the context branch, which models short-range context among pixels, resulting in better representation learning. Moreover, the task similarity constraint leads to long-range context invariance, enabling MaskSup to better learn the shape of the ROI, which in turn translates into better output predictions (see Figure~\ref{Fig:qualmask} for more comparative results). Overall, learning with the context branch and task similarity constraint helps in cases of segmenting ambiguous ROIs at varying size and color, and also better segment minority class instances in cases of multi-class segmentation.

\subsection{Ablation study}
\begin{figure}[!htb]
\centering
\setlength{\tabcolsep}{5pt}
\begin{tabular}{ccc}
\includegraphics[width=1.55in]{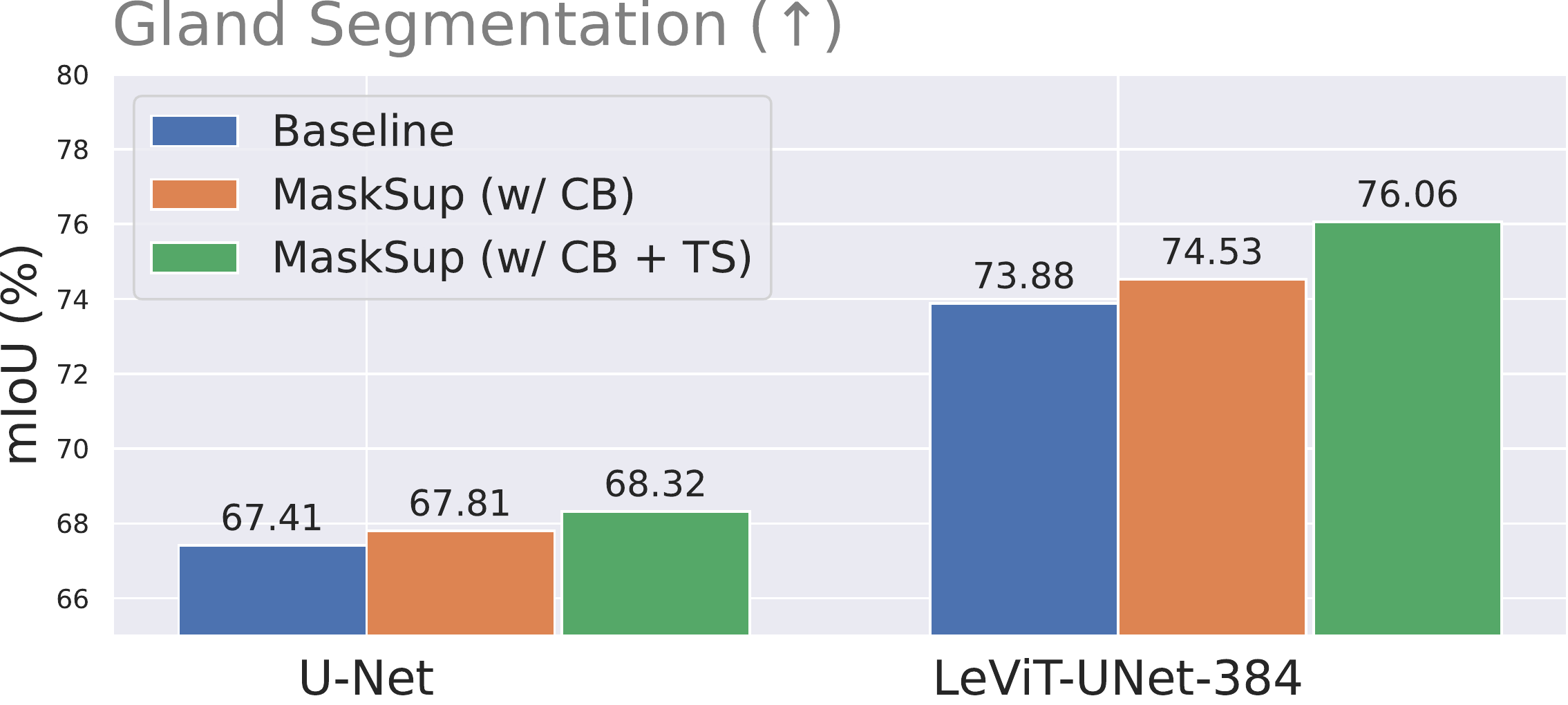} & \includegraphics[width=1.55in]{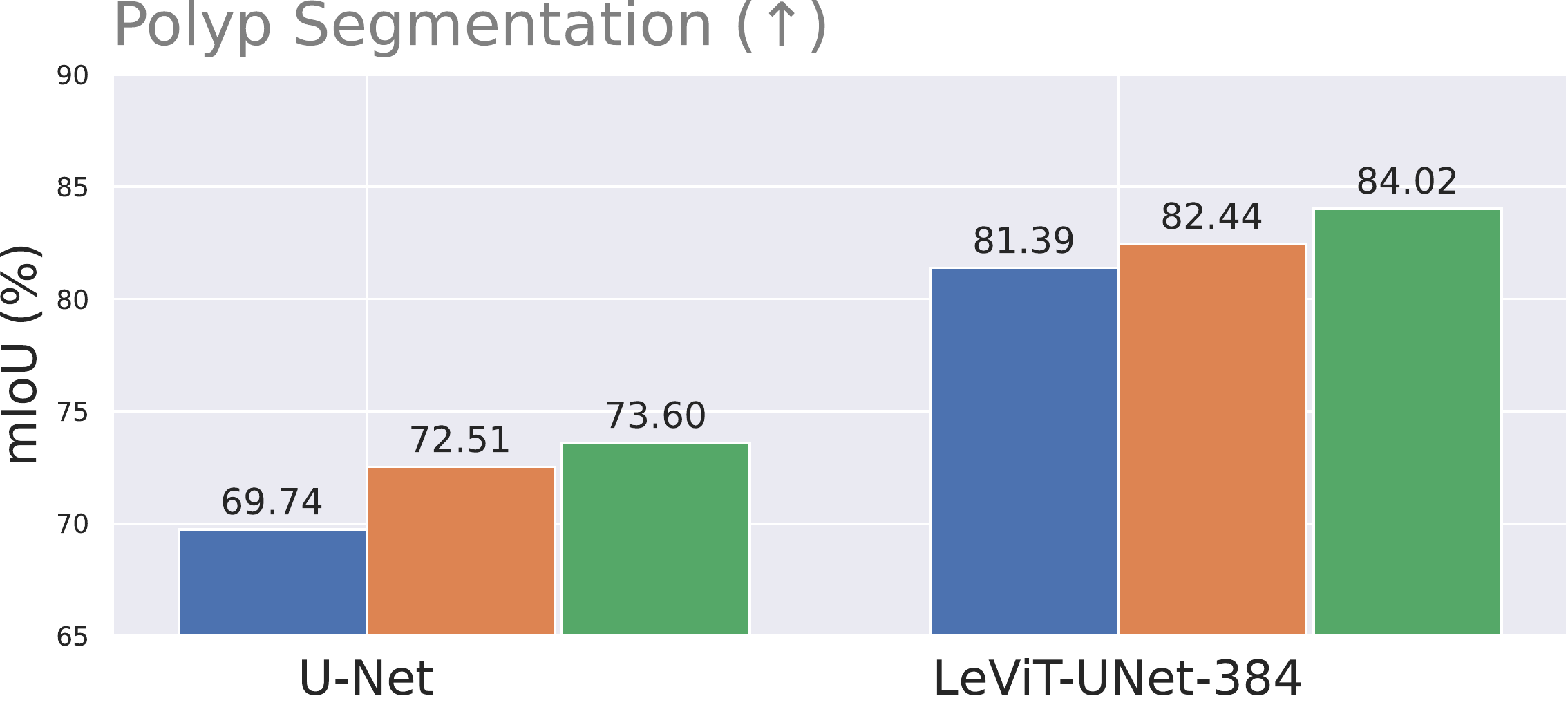} & \includegraphics[width=1.55in]{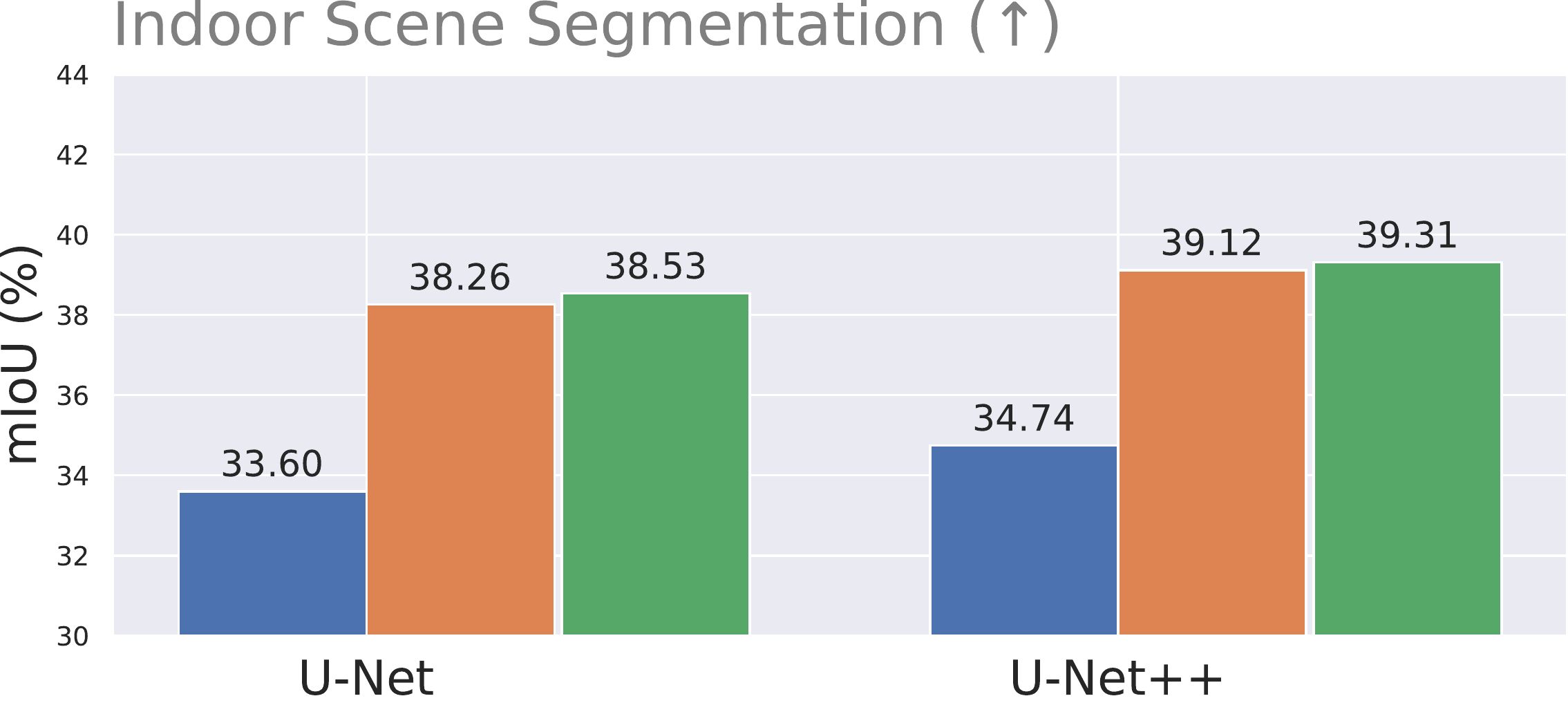}
\end{tabular}
\caption{Ablation study of different modules of MaskSup on GLaS, Kvasir \& CVC-ClinicDB and NYU Depth V2 test sets. MaskSup (CB \& TS) consistently improves performance of various baselines in both binary and multi-class image segmentation tasks.}
\label{Fig:ablation}
\end{figure}

\noindent\textbf{Effectiveness of Context Branch.}\quad Figure~\ref{Fig:ablation} illustrates the benefit of using the the context branch on both convolutional and transformer-based methods. Using the the context branch leads to better modeling of local semantics, as it is tasked to output pixel-wise predictions for \textbf{masked regions} in the input; thereby leveraging information from neighboring pixels. We can see that the context branch improves performance of different segmentation methods across the three datasets. This shows that MaskSup is generic and can be easily integrated into existing image segmentation methods. However, it is important to mention that a higher performance improvement is observed when the architecture is a transformer-based method due to its key characteristic of modeling long-range context~\cite{dosovitskiy2020vit,xu2021levit}.

\medskip\noindent\textbf{Effectiveness of Task Similarity Constraint.}\quad Figure~\ref{Fig:ablation} shows the benefit of using the task similarity constraint. We observe that the task similarity constraint further improves performance of both convolutional and transformer-based methods. The use of the task similarity constraint results in learning long-range context invariant representations, which help capture the shape of the ROI. This, in turn, leads to accurate predictions even in cases of different shapes and appearance of instances, ambiguous ROIs at different sizes, and imbalance among multiple class instances in multi-class segmentation.

\medskip\noindent\textbf{Amount of Masking.}\quad We performed an ablation study of high- and low-masked pixels regions during MaskSup training, and the results are reported in Table~\ref{Tab:mask}, which shows that masking the images heavily during training yields better performance of MaskSup across all three datasets.

\begin{table}[!htb]
\caption{Ablation study of high- and low-masked pixels regions during MaskSup training.}
\medskip
\centering
\begin{tabular}{l*{7}{c}}
\toprule
\textbf{Masking} & \textbf{GLaS, mIoU ($\uparrow$)} & \textbf{CVC-Clinic-DB, mIoU ($\uparrow$)} & \textbf{NYUDv2 ($\uparrow$)} \\
\midrule
Low & 75.65 & 81.80 & 35.33 \\
High & \textbf{76.06} & \textbf{84.02} & \textbf{39.31} \\
\bottomrule
\end{tabular}
\label{Tab:mask}
\end{table}

\subsection{Analysis}
\begin{figure}[!htb]
\centering
\includegraphics[scale=.28]{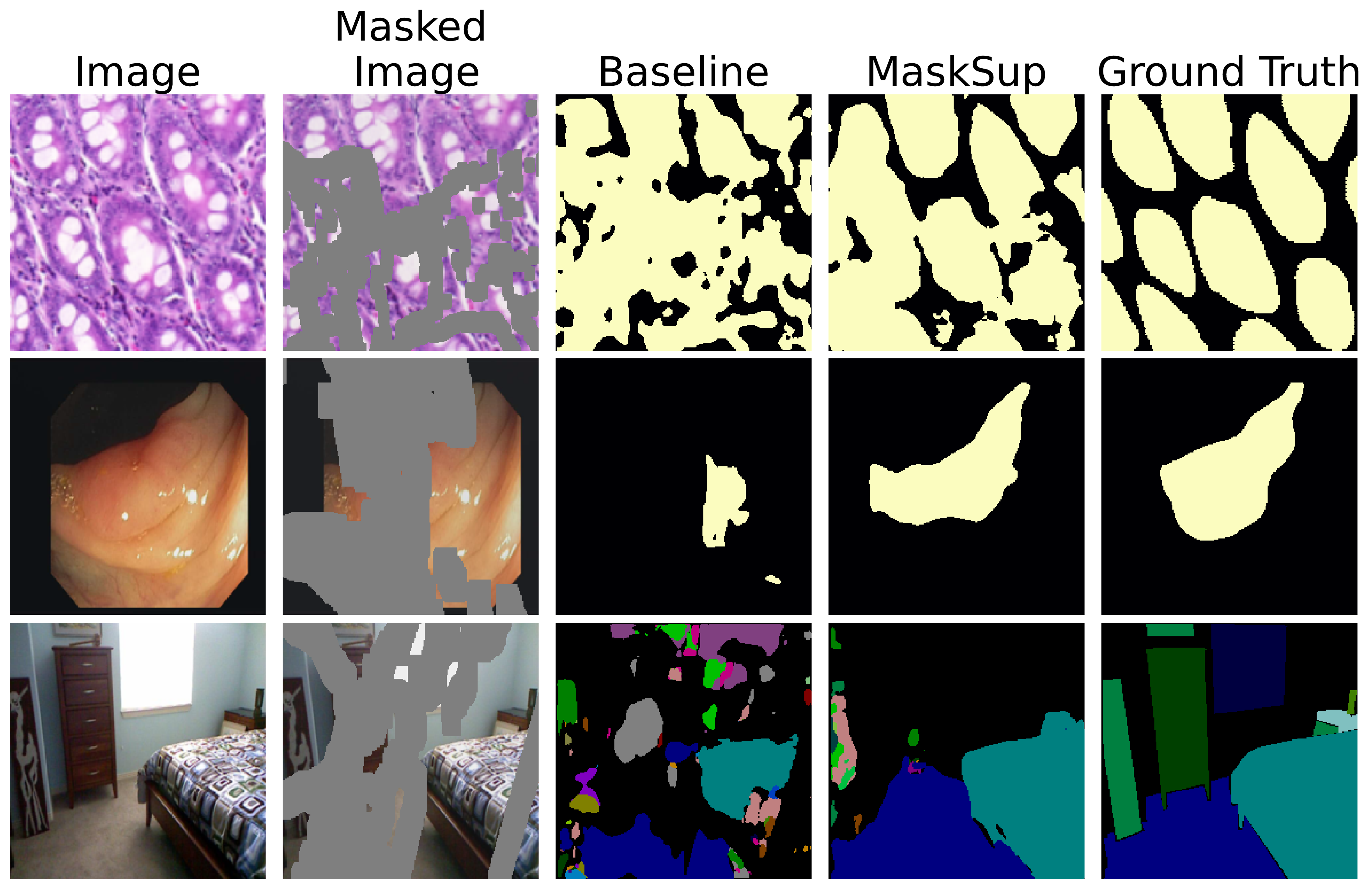}
\caption{Visual comparison of predictions made by MaskSup and baseline for images with \textbf{masked} regions.}
\label{Fig:qualmask}
\end{figure}

\noindent\textbf{Robustness to Masked Corruptions.}\quad Figure~\ref{Fig:qualmask} shows the robustness of MaskSup to masked corruptions. As can be seen, MaskSup is able to better predict the ROI even when a large portion of the image is masked, demonstrating its capability in modeling short- and long-range context. Using both the context branch and task similarity constraint, MaskSup is able to learn context invariant representations to better segment and preserve the ROI shape.

\medskip\noindent\textbf{Computational Efficiency.}\quad In Table~\ref{Tab:scale}, we report the number of parameters in millions (M), as well as mIoU for MaskSup and baseline methods. MaskSup with LeViT-192 network outperforms LeViT-384~\cite{xu2021levit} on both GLaS and Kvasir \& CVC-ClinicDB, while having almost $2.6 \times$ fewer learnable parameters. MaskSup with U-Net also outperforms U-Net++~\cite{zhou2019unet++}, which has almost $3\times$ more learnable parameters on NYUDv2, with a relative improvement of 10.91\% in terms of mIoU. Hence, there is no trade-off between segmentation accuracy and computational efficiency when using MaskSup in comparison with scaled versions of the networks.

\begin{table}[!htb]
\caption{Comparison of MaskSup and baselines on GLaS, Kvasir \& CVC-ClinicDB and NYUDv2 test sets. MaskSup is computationally efficient and achieves superior performance with fewer parameters. Boldface numbers indicate better performance.}
\small
\setlength{\tabcolsep}{2.2pt}
\medskip
\centering
\begin{tabular}{l*{7}{c}}
\toprule
\textbf{Method} & Params (M) ($\downarrow$) & GLaS, mIoU ($\uparrow$) & CVC-Clinic-DB, mIoU ($\uparrow$) & NYUDv2 ($\uparrow$) \\
\midrule
LeViT-384~\cite{xu2021levit} & 51 & 73.88 & 81.38 & - \\
MaskSup (LeViT-192)  & \textbf{19}\textcolor{blue}{(2.6x)} & \textbf{74.44}\textcolor{blue}{(+0.75)} & \textbf{82.17}\textcolor{blue}{(+0.97)} & - \\
\midrule
U-Net++~\cite{zhou2019unet++} & 9 & - & - & 34.74 \\
MaskSup (U-Net) & \textbf{3}\textcolor{blue}{(3x)} & - & - & \textbf{38.54}\textcolor{blue}{(+10.91)}\\
\bottomrule
\end{tabular}
\label{Tab:scale}
\end{table}

\section{Conclusion}
We introduced a new learning paradigm, called Masked Supervised Learning, for semantic segmentation. By constructing a randomly masked version of the input image, we first make predictions using a base segmentation network on the two inputs. Then, we maximize the predictions between the two outputs to model both short- and long-range context. MaskSup can be easily integrated into any existing semantic segmentation method. We show that MaskSup achieves better performance than state-of-the-art single and multi-task learning baselines in both binary and multi-class semantic segmentation tasks, especially in tackling small, ambiguous regions and minority class instances. In addition, MaskSup is robust to masked corruptions and is computationally efficient without compromising performance.

For future work, we aim to investigate what type of masking strategies works best in MaskSup. Since MaskSup is a generic paradigm, we plan to adapt it to other computer vision tasks such as multi-label recognition, object detection and human pose estimation. We also plan to explore high-resolution segmentation (e.g. Cityscapes, ADE20K datasets) using MaskSup.

\bibliography{references}
\end{document}